\PassOptionsToPackage{table}{xcolor}

\documentclass[sigconf]{aamas}

\usepackage{balance} 

\setcopyright{none}
\acmConference[ALA '26]%
{Proc.\@ of the Adaptive and Learning Agents Workshop (ALA 2026)}%
{May 25 -- 26, 2026}%
{Paphos, Cyprus, https://alaworkshop2026.github.io/}%
{Aydeniz, Delgrange, Mohammedalamen, Yang (eds.)}%
\copyrightyear{2026}
\acmYear{2026}
\acmDOI{}
\acmPrice{}
\acmISBN{}
\usepackage[smaller]{acronym}
\usepackage{xspace}
\usepackage{amsmath}
\usepackage{amsthm}
\usepackage{subcaption}

\usepackage{footnote}
\makesavenoteenv{table}

\usepackage{array}
\usepackage{multirow}
\usepackage{booktabs}

\usepackage{tikz}
\tikzset{
    agent/.style={
        thick, fill=blue!30, draw,
        rectangle, minimum width=0.4cm, minimum height=0.4cm
    },
    agent-head/.style={
        thick, fill=blue!50, draw,
        rectangle, minimum width=0.3cm, minimum height=0.15cm
    }
}

\usepackage{footmisc}

\acrodef{SAM}{Safe Action Model Learning}
\acrodef{PAC}{Probably Approximately Correct}
\acrodef{NSAM}{Numeric Safe Action Model Learning}
\newtheorem{definition}{Definition}

\newcommand{\nsam}{NSAM\xspace}

\acrodef{RL}{Reinforcement Learning}

\newcommand{\rl}{\ac{RL}\xspace}
\acrodef{IL}{Imitation Learning}

\acrodef{BC}{Behavioral Cloning}

\acrodef{GAIL}{Generative Adversarial Imitation Learning}

\acrodef{PPO}{Proximal Policy Optimization}
\newcommand{\ppo}{\ac{PPO}\xspace}

\acrodef{RAMP}{Reinforcement learning, Action Model learning, and Planning}
\newcommand{\hybrid}{\ac{RAMP}\xspace}

\acrodef{DQN}{Deep Q-Network}

\acrodef{QR-DQN}{Quantile Regression DQN}

\newcommand{\mcraft}{Minecraft\xspace}

\newif\ifaddcomments

\newcommand{\commentout}[1]{ }
\newcommand{\miniparagraph}[1]{\noindent\textbf{#1 .}}

\newcommand{\yarin}[1]{\ifaddcomments{\textcolor{violet}{[Yarin: #1]}}\fi}

\acmSubmissionID{33}

\title{RAMP: Hybrid DRL for Online Learning of Numeric Action Models}

\author{Yarin Benyamin}
\affiliation{
  \institution{Ben-Gurion University of the Negev}
  \city{Be'er Sheva}
  \country{Israel}}
\email{bnyamin@post.bgu.ac.il}

\author{Argaman Mordoch}
\affiliation{
  \institution{Ben-Gurion University of the Negev}
  \city{Be'er Sheva}
  \country{Israel}}
\email{mordocha@post.bgu.ac.il}

\author{Shahaf Shperberg}
\affiliation{
  \institution{Ben-Gurion University of the Negev}
  \city{Be'er Sheva}
  \country{Israel}}
\email{shperbsh@bgu.ac.il}

\author{Roni Stern}
\affiliation{
  \institution{Ben-Gurion University of the Negev}
  \city{Be'er Sheva}
  \country{Israel}}
\email{roni.stern@gmail.com}

\begin{abstract}
Automated planning algorithms require an \emph{action model} specifying the preconditions and effects of each action, but obtaining such a model is often hard. Learning action models from observations is feasible, but existing algorithms for numeric domains are offline, requiring expert traces as input.
We propose the \hybrid strategy for learning numeric planning action models online via interactions with the environment. 
\hybrid simultaneously trains a Deep Reinforcement Learning (DRL) policy, learns a numeric action model from past interactions, 
and uses that model to plan future actions when possible. 
These components form a positive feedback loop: the RL policy gathers data to refine the action model, while the planner generates plans to continue training the RL policy. To facilitate this integration of RL and numeric planning, we developed Numeric PDDLGym, an automated framework for converting numeric planning problems to Gym environments.
Experimental results on standard IPC numeric domains show that \hybrid significantly outperforms PPO, a well-known DRL algorithm, in terms of solvability and plan quality.
\end{abstract}

\keywords{Deep Reinforcement Learning; Numeric Planning; Action Model Learning; Online Learning;}

\newcommand{\BibTeX}{\rm B\kern-.05em{\sc i\kern-.025em b}\kern-.08em\TeX}

\begin{document}


\pagestyle{fancy}
\fancyhead{}


\maketitle 


\section{Introduction}

Automated planning~\citep{ghallab2004automated} is a reliable approach for sequential decision-making, but it requires an \emph{action model} specifying the preconditions and effects of each action. 
Handcrafting action models is notoriously difficult, especially for \emph{numeric planning problems}, where action preconditions and effects involve discrete and continuous state variables. 
Prior work proposed automated methods for action model learning (AML) in numeric planning~\cite{mordoch2023learning,segura2021discovering} were \emph{offline}, requiring experts to provide execution traces to learn from. 
We consider the \emph{online learning} setting, where an agent must learn from its own interactions with the environment. 
Deep Reinforcement Learning (DRL) algorithms such as \ppo~\cite{schulman2017proximal} are designed for such online learning settings, but they often lack the structural advantages of symbolic planning.

In this work, we present \hybrid, a novel strategy for solving numeric planning problems that integrates DRL, online AML, and planning. \hybrid executes a positive feedback loop: an RL policy is used to explore the environment and collect data in a goal-oriented manner. This data is used to learn a numeric planning action model. This learned model is then utilized by a planner to generate high-quality plans, which in turn accelerate the training of the DRL policy.
To enable this DRL-planning integration, we developed a framework for automatically converting numeric planning domains specified in PDDL 2.1~\citep{fox2003pddl2} into standard Gym environments~\citep{towers2024gymnasium}. 

We implemented \hybrid and evaluated empirically on three domains from the International Planning Competition (IPC) and a recently proposed numeric planning domain inspired by \mcraft~\citep{benyamin2024crafting}. 
Results show that \hybrid is able to learn effective action models and use them to solve more problems and find better solutions than PPO~\cite{schulman2017proximal}, a state-of-the-art DRL baseline.

\section{Background}
\label{sec:background}
A \emph{numeric planning domain} in PDDL2.1~\citep{fox2003pddl2} is defined by a tuple $D=(A, T, F, X, P)$, where $A$ is a set of actions, 
$T$ is a set of object types, 
$F$ is a set of fluents, and $X$ is the set of functions. A fluent is a Boolean variable, while a function is a numeric variable. 
Every action in $A$ is associated with a set of \emph{preconditions} (logical and linear numeric constraints) and \emph{effects} (assignments or modifications to the fluents and functions). 
Domains are usually defined in a \emph{lifted} manner, which means that the fluents, functions, and actions are parameterized by object types. 
A \emph{numeric planning problem} in a domain $D$ is defined by a tuple $P=(s_0, O, G)$, where $s_0$ is an initial state, $O$ is a set of objects, and $G$ is a goal definition. 
A grounded fluent is a fluent with specific objects in $O$ assigned to its parameters.
Grounded functions and actions are similarly defined. 
A state in a numeric planning problem is an assignment of values to all the grounded fluents and functions. 
A goal in numeric planning is a set of constraints over the grounded fluents and functions, and a goal state is any state that satisfies the goal constraints. 
A solution to a planning problem is a \emph{plan}, which is an applicable sequence of actions leading from an initial state to a goal state.

\miniparagraph{Learning Domain Models}
\yarin{This part got shortened and moved to the related work}
Automated planning algorithms assume the existence of a symbolic model. However, obtaining such a model is challenging, motivating the development of automated Action Model Learning (AML) algorithms. 
Research has primarily focused on the \emph{offline} setting, where the AML algorithm is given a set of \emph{trajectories} to learn from. A trajectory is an alternating sequence of states and actions, where the state after an action is the result of applying the action on the state immediately before it. 

In the \emph{online} learning setting, no expert trajectories are given. Instead, an agent interacts with the environment in a sequence of episodes to maximize cumulative reward or minimize plan length. 
While several online learning algorithms have been proposed for classical planning domains, there are currently no methods for online learning of numeric planning domains. 

A critical property of some AML algorithms is that they provide a form of ``safety'' guarantee: not only must the learned model be consistent with the given observations, but every plan generated from it is guaranteed to be \emph{sound} with respect to the true, unknown action model.

\begin{definition}[Safe Domain Model]\label{def:safe}
Let $M^*$ be the true, unknown action model of an environment, and let $M$ be a learned domain model. $M$ is considered \emph{safe} if every plan that is valid in $M$ is \emph{sound} with respect to $M^*$. Therefore, for any initial state $s_0$ and goal $G$, if a plan $\pi = \langle a_1, a_2, \dots, a_n \rangle$ reaches $G$ under the transition dynamics of $M$, it is fully executable under $M^*$ from $s_0$, and its execution in $M^*$ results in a state $s_n$ that satisfies $G$.
\end{definition}

\miniparagraph{Reinforcement Learning (RL)}
Reinforcement Learning (RL)~\citep{sutton2018reinforcement} is a framework for solving decision-making problems where an agent learns a policy $\pi(a|s)$---mapping a state $s$ to an action $a$---to maximize expected cumulative rewards. Deep RL (DRL) extends this by training neural networks to represent the policy. 
A notable RL algorithm, which we make use of in this work, is Proximal Policy Optimization (PPO)~\citep{schulman2017proximal}, a stable policy-gradient algorithm that provides state-of-the-art performance in discrete decision-making environments. It operates by alternating between sampling data through environmental interaction and optimizing a surrogate objective function, utilizing a clipping mechanism to prevent destructively large policy updates and ensure stable learning.

Standard DRL algorithms are known to struggle in planning problems that require reasoning over long horizons.
Also, most DRL algorithms are designed for richer environments than numeric planning, including stochastic effects and raw observations such as images. Thus, they are expected to perform poorly on problems that can be adequately abstracted as symbolic numeric planning problems. 
Hybrid RL-symbolic approaches like SK~\citep{sreedharan2023optimistic} and SORL~\citep{jin2022creativity} have been proposed, but do not address symbolic numeric planning.

Our work addresses this gap by introducing \hybrid, the first strategy for \emph{online} learning of \emph{numeric} action models, integrating safe model learning with Deep RL.

\noindent\textbf{Problem Setting.}
The agent is situated in an environment modeled as a numeric planning domain $D$.
The agent has access to the domain’s types, fluents, functions, and action names, but not the domain's action model, i.e., it does not know the actions' preconditions and effects.
In every \emph{episode}, a problem $p=(s,O,G)$ in this domain is given to solve. 
The agent then chooses which action to perform, leading to a new state $s'$. 
This process is repeated until the problem is solved, i.e., $s'$ satisfies the goal $G$, or until reaching a maximum number of steps $t_{MAX}$. 
The objective is to learn a policy $\pi$ that solves problems efficiently, i.e., that guides the agent to a goal with as few steps as possible.

\section{Related Work}\label{sec:related-work}
In this section, we discuss several lines of research that are related to this work. 

\subsection{Offline Action Model Learning Algorithms}
\label{sec:related-offline-action-model-learning}

FAMA~\citep{aineto2019learning} is an offline action model learning algorithm that can handle missing observations and outputs a classical planning domain model. 
It frames the task of learning an action model as a planning problem, ensuring that the returned action model is consistent with the provided observations.
NOLAM~\citep{Lamanna24} can learn action models even from noisy trajectories. 
LOCM~\citep{cresswell2013acquiring} learns an action model from observed sequences of actions and their signatures, without observing the states in the trajectory.

The Safe Action Model (SAM) learning algorithm~\cite{stern2017efficient,juba2021safe} differs from the above algorithm in that the action model it returns provides a form of ``safety'' guarantee (Definition~\ref{def:safe}).
SAM has been extended to support lifted action model representation~\citep{juba2021safe}, partial observability~\citep{le2024learning}, stochastic effects~\citep{juba2022learning}, and conditional effects~\cite{MordochSSJ24}. 
All SAM algorithms require observing all the actions in the given trajectories, and most also require observing the states. 
\emph{NSAM}~\citep{mordoch2023learning}, the algorithm we use in this work, extends SAM to numeric domains while preserving the same safety guarantees.

\yarin{New:}
While the aforementioned algorithms rely on structured execution traces, other approaches extract models from less structured knowledge. 
Framer~\citep{lindsay2017framer} induces planning models by clustering semantic frames in textual activity descriptions. 
In contrast, Large Language Models (LLMs) leverage pre-trained knowledge to construct models from language and can support interactive planning workflows for execution and validation~\cite{tantakoun2025llms,benyamin2025toward}.
While these approaches reduce the barrier to model creation, they do not address online learning of numeric action dynamics.

To handle environments where neither symbolic nor language inputs are available, recent work has explored learning action models directly from raw images. 
LatPlan~\citep{asai2018classical} learns propositional action models in the latent space using a variational autoencoder. 
They use the Gumbel-Softmax technique~\citep{jang2017categorical} to convert the continuous output of an autoencoder into categorical variables. 
These categorical variables are used as propositional symbols in a symbolic reasoning system, which, in LatPLan's case, is a symbolic action model. 
ROSAME-I~\citep{xi2024neuro}, like LatPlan, learns action models from visual inputs. 
Unlike LatPlan, ROSAME-I requires knowing the set of possible propositions and action signatures as input.
ROSAME-I simultaneously learns classifiers for identifying propositions in a given image and infers a lifted, first-order action model defined over the given set of propositions and actions.

Although these approaches handle symbolic or unstructured inputs, they do not specifically address numeric planning problems. 
\nsam~\cite{mordoch2023learning} and 
PlanMiner~\citep{segura2021discovering} are, to the best of our knowledge, the only algorithms capable of learning action models that include both discrete and numeric preconditions and effects.\footnote{Some other action model learning algorithms are capable of learning the numeric costs or rewards associated with actions~\cite{jin2022creativity}.}  
\nsam makes several simplifying assumptions, full, noise-free observability, conjunctions of linear inequalities for preconditions, and conjunctions of linear equations for effects, which allow it to run in polynomial time. Under these assumptions, it guarantees a \emph{safe domain model} by computing the minimal numeric preconditions via convex hulls over successful states and exact numeric effects through linear equation solving. In contrast, PlanMiner can handle noisy observations but does not provide safety guarantees and requires solving a generally intractable symbolic regression problem. For these reasons, we selected \nsam as our primary action model learning algorithm.

\subsection{Online Action Model Learning Algorithms}
\label{sec:related-online-action-model-learning}
Online action-model learning algorithms 
iteratively learn an incumbent action model and choose the next actions to perform in order to collect observations that enable further refinement of the incumbent action model. 
OLAM~\citep{lamanna2021online} is an online action model learning algorithm that is designed for classical planning domains. It identifies in every iteration an action and a state where trying to execute that action is expected to refine the incumbent action model. 
Then, it uses a planner to find a plan to reach that state and attempts to execute the chosen action. 
GLIB~\citep{chitnis2021glib} follows a similar approach but is designed for stochastic environments, resulting in a Probabilistic PDDL (PPDDL)~\citep{younes2004ppddl1} action model.
QACE~\citep{verma2023autonomous} is an action model learning algorithm that can also query a black-box expert. It outputs a PPDDL action model with the same capabilities as the black-box expert it trained from. 
Karia et al.~\citeyear{karia2023epistemic} extend QACE to address the non-stationarity of the environment, i.e., address cases where the environment dynamics change. QACE+ achieves this by interleaving planning and learning and focusing on learning only the models essential for the tasks at hand.
ILM~\citep{ng2019incremental} employs an explore-exploit strategy: if it reaches a state from which the goal can be achieved, it exploits this state; otherwise, it explores through random walks.
Instead of focusing solely on reaching a specific goal, the agent can take a broader approach by exploring the environment and aiming for an interesting state in it.

Recent works explored integrating \rl and online learning of a symbolic action model~\cite{jin2022creativity,sreedharan2023optimistic}. 
The objective of these works is typically to maximize a cumulative reward metric, in contrast to action model learning algorithms like OLAM and ILM, whose objective is to learn a symbolic action model.
Sreedharan and Katz~\cite{sreedharan2023optimistic} proposed such an algorithm, which we refer to as the SK algorithm. SK begins by initializing an optimistic symbolic model that assumes all actions are applicable in every state (i.e., no preconditions) and the effect of each action includes all grounded predicates. It then employs fast, diverse planners to generate potential paths toward the goal. While these paths are unlikely to be valid, they serve as exploration mechanisms to gather new information. 
Specifically, these plans are executed within the environment, with the outcomes used to train a reward-maximizing policy using Q-learning~\citep{watkins1989learning,watkins1992q}.
This continuous process of exploration and symbolic model refinement is guaranteed to generate a goal-reaching policy. Our hybrid strategy for the online learning setting is somewhat similar to SK. However, SK is only designed for classical (non-numeric) planning, and its applicability to numeric planning remains uncertain. 
SORL~\citep{jin2022creativity} is another online algorithm that integrates RL and learning symbolic action models to maximize the cumulative reward. 
It collects visual observations from the environment and assumes the existence of a mapping function from visual observations to symbolic states. 
SORL iteratively learns and creates 
a symbolic, higher-level action model, 
and a lower-level set of RL policies, referred to as \emph{symbolic options}. 
A planner uses the learned symbolic action model to create a high-level plan, and a meta-controller chooses or creates symbolic options to try to execute the high-level plan, exploring the environment as needed.

While not explicitly specified, the action model learning algorithm SORL uses is not robust to missing or noisy observations. It assumes that an action's effects can be inferred by the difference between the states observed before and after applying that action and shows no support for numeric preconditions. Numeric effects are supported in a very limited way, only learning which actions increase the reward and by how much. Other numeric aspects, e.g., numeric state variables and preconditions, are not learned.


\subsection{Summary: Action Model Learning Algorithms}
\label{sec:summary-action-model-learning}

\begin{table}[ht]
\caption{Comparison of various action model learning algorithms, based on their support of given problem - numeric inputs, stochasticity, non-stationarity, noisy, observability, and online/offline learning capability.}
\centering
\resizebox{\columnwidth}{!}{
\begin{tabular}{|l|c|c|c|c|c|c|c|}
\hline
\textbf{} & \textbf{Input} & \textbf{Numeric} & \textbf{Stochastic} & \textbf{NS} & \textbf{Noisy} & \textbf{Obs.} & \textbf{Online/Offline} \\
\hline
\textbf{FAMA~\citep{aineto2019learning}} & symbolic & No & No & No & No & Partial & Offline \\
\hline
\textbf{LOCM~\citep{cresswell2013acquiring}} & symbolic & No & No & No & No & Only action & Offline \\
\hline
\textbf{Framer~\citep{lindsay2017framer}} & text & No & No & No & - & Only action & Offline \\
\hline
\textbf{OLAM~\citep{lamanna2021online}} & symbolic & No & No & No & No & Yes & Online \\
\hline
\textbf{NOLAM~\citep{Lamanna24}} & symbolic & No & No & No & Yes & Yes & Offline \\
\hline
\textbf{ILM~\citep{ng2019incremental}} & symbolic & No & Yes & Yes & Yes & Yes & Online \\
\hline
\textbf{GLIB~\citep{chitnis2021glib}} & symbolic & No & Yes & No & No & Yes & Online \\
\hline
\textbf{QACE~\citep{verma2023autonomous}} & symbolic & No & Yes & No & No & Yes & Online \\
\hline
\textbf{QACE+~\citep{karia2023epistemic}} & symbolic & No & Yes & Yes & No & Yes & Online \\
\hline
\textbf{SORL~\citep{jin2022creativity}} & visual\footnote{They assumed as input a perfect mapping from visual input to symbolic state.}  & No\footnote{The support for numeric planning is limited to only learning how much reward each action adds.} & No & No & - & Yes & Online \\
\hline
\textbf{SAM~\citep{juba2021safe}} & symbolic & No & No & No & No & Yes & Offline \\
\hline
\textbf{NSAM~\citep{mordoch2023learning}} & symbolic & Yes & No & No & No & Yes & Offline \\
\hline
\textbf{PlanMiner~\citep{segura2021discovering}} & symbolic & Yes & No & No & Yes & Partial & Offline \\
\hline
\textbf{SK~\citep{sreedharan2023optimistic}} & symbolic & No & No & No & No & Yes & Online \\
\hline
\textbf{JRK~\citep{james2022autonomous}} & visual & No & Yes\footnote{While they learn a PPDDL model, the experimental results all use a deterministic planner.} & No & - & Yes & Offline \\
\hline
\textbf{LATPLAN~\citep{asai2018classical}} & visual & No & No & No & - & Yes & Offline \\
\hline
\textbf{ROSAME-I~\citep{xi2024neuro}} & visual & No & No & No & - & No\footnote{Note that the first and last states in every trajectory are assumed to be fully obseravable} & Offline \\
\hline
\rowcolor{yellow}
\textbf{\hybrid (our method)} & symbolic & Yes & No & No & No & Yes & Online  \\
\hline
\end{tabular}
}
\label{tab:methods-comparison}
\end{table}

Table~\ref{tab:methods-comparison} provides an overview of all action model learning algorithms described above. 
Every row represents a model-learning algorithm, and every column represents a property of action model-learning algorithms. 
Column ``Input'' refers to the type of input given to the learning algorithm, namely, whether it is symbolic, text or visual. 
Columns ``Numeric'' and ``Stochastic'' refer to whether the underlying environment includes numeric state variables and stochastic effects, respectively. 
Column ``NS'' (non-stationarity) 
refers to whether the dynamics of the underlying environment, i.e., the actions' preconditions and effects, may change during learning. 
Columns ``Noisy'' and ``Obs.'' refer to whether the states and actions in the given observations are noisy and fully observed, respectively. Note that if the ``Input'' is visual, the algorithm can handle noise due to its use of function approximation for image processing. This is indicated by ``-'' in Table~\ref{tab:methods-comparison}. 
The ``Online/Offline'' column refers to whether the learning algorithm is an online algorithm or an offline algorithm. 
As can be seen, the \hybrid strategy we propose in this work is the only online learning algorithm that supports learning a numeric action model.

\section{The RAMP Strategy}
\label{sec:hybrid}

\begin{figure}
    \centering
    \includegraphics[width=0.7\linewidth]{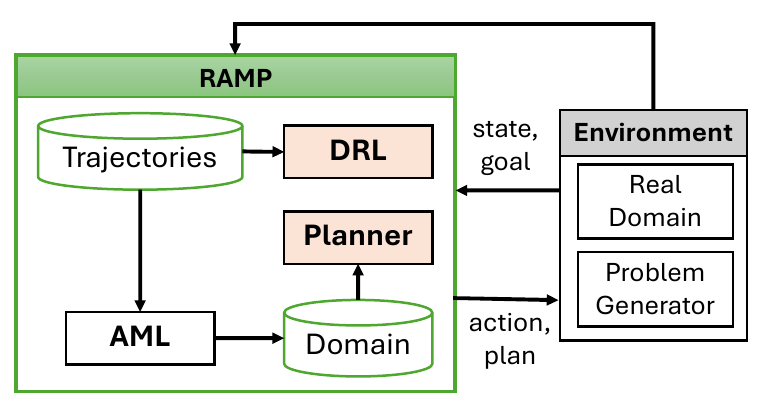}
    \caption{A high-level diagram of the \hybrid strategy.}
    \label{fig:ramp-diagram}
\end{figure}

\hybrid integrates three components: a DRL algorithm, an AML algorithm, and a numeric planner. 
It maintains a set $\mathcal{T}$ of observed trajectories 
and an incumbent learned domain model, denoted $M$. 
Both are initialized to be empty. 
At the beginning of every episode, \hybrid attempts to find a plan with the planner using $M$.\footnote{In the first iteration, $M$ is empty, so the planner cannot run; we treat this as a failed planning attempt.}  
If a plan is found, the agent executes this plan.
Otherwise, \hybrid uses the DRL algorithm to choose which actions to perform in this episode. 
At the end of every episode, we add the resulting trajectory to $\mathcal{T}$. 
This trajectory is used as a training example for both the DRL algorithm and the AML algorithm. 
\yarin{added:} Re-running the offline AML on the cumulative trajectory set at the end of each episode is what effectively turns it into an online method.
Figure~\ref{fig:ramp-diagram} provides a high-level illustration of \hybrid.

The integration of AML, planning, and DRL in \hybrid establishes a symbiotic relationship between them. 
For the planner, the DRL algorithm acts as a fail-safe mechanism when it fails to find a plan.
For the AML algorithm, the DRL  algorithm provides a principled method for gathering observations in a goal-oriented manner, leveraging its inherent ability to balance exploration and exploitation. 
Simultaneously, the DRL algorithm uses the trajectories created by the planner for training. 
These trajectories tend to represent efficient ways to solve the problem. 
Such high-quality data improves sample efficiency and stabilizes the learning process, as we observed empirically in our experimental results.

\miniparagraph{Implementation Details}
While \hybrid is agnostic to the AML and DRL algorithms used, in our implementation, we use \nsam~\footnote{Technically speaking, our agent uses \nsam*~\cite{mordoch2024-nsam-star}, an advanced version of \nsam.} for AML and the RLlib implementation of \ppo for DRL.
\nsam is one of only two existing AML algorithms that can learn numeric domains, and was chosen for its ability to return safe action models (Definition~\ref{def:safe}). 
\ppo was used in our implementation of \hybrid since it is a well-known, popular, stable, and has low sensitivity to hyperparameters.
We have also considered alternative DRL algorithms such as Rainbow DQN~\citep{hessel2018rainbow} and Soft Actor-Critic (SAC)~\citep{haarnoja2018soft}, but they consistently failed to learn across these continuous environments despite extensive hyperparameter tuning.

A challenge when using \ppo is its \emph{clipping mechanism}, which is designed to constrain policy updates by preventing the probability ratio between the new and old policies from deviating too far from 1. When the agent follows the plan generated by the planner, its own learned policy may assign a low probability to the dictated actions, creating a significant discrepancy between the executed actions and those preferred by the current policy. This can cause the importance-weighted policy ratio to frequently fall outside the clipping range, effectively nullifying the gradient update and slowing down learning. 
As a result, in some cases, \ppo may struggle to meaningfully adjust its policy, particularly if the expert actions are substantially different from what it would naturally choose. 
To address this issue, we adopt an approach for masking invalid actions~\citep{HuangO22}. Specifically, we treat the expert action in each state as the only valid action and mask out all others. This ensures that the logits of actions that do not conform to the plan are set to zero, preventing them from influencing the policy update. As a result, the gradient for these actions is not eliminated, ensuring that the update is not affected by PPO's clipping mechanism.

Similarly, off-the-shelf DRL algorithms such as \ppo are designed for stochastic environments. 
Thus, the agent may repeatedly attempt the same inapplicable action in a given state. To prevent this, we apply a \emph{masking} mechanism that disallows actions previously observed to be inapplicable.
Similar masking techniques have been used in DRL in other domains~\cite{vinyals2017starcraft,berner2019dota,ye2020mastering}.

\section{Automated PDDL to Gym Conversion}
\label{sec:pddl-to-gym}

RAMP requires the use of DRL algorithms to solve numeric PDDL planning problems. 
DRL algorithms work by interacting with the \emph{environment}, which in our case is given by the PDDL problem. 
PDDLGym~\citep{silver2020pddlgym} provides a wrapper over a selected number of PDDL problems that allows simulating ``interactions'' with them and using RL algorithms to attempt to solve them. 
Specifically, they provide AI Gym~\citep{towers2024gymnasium} environments that simulate classical PDDL domains. 
This is particularly useful since standard implementations of RL algorithms, such as RLlib~\citep{liang2018rllib} and Stable Baselines~\cite{stable-baselines3}, support the Gym interface.  
However, PDDLGym's design is primarily tailored to tabular RL methods and does not support numeric planning. 

Therefore, we developed an automated framework to convert PDDL2.1 domains into AI Gym environments, which we refer to as Numeric PDDLGym.\footnote{The source code for the Numeric PDDLGym environment is available at: https://github.com/SPL-BGU/NumericPDDLGym}
This framework provides full support for Boolean and numeric state variables and lifted domain representation. 
It accepts a PDDL domain and problem file as input and generates a Gym environment that simulates the corresponding planning problem. 
A challenge in this conversion mechanism is that standard RL and DRL algorithms expect fixed-size observation and action spaces, i.e., a fixed number of state variables and actions. 
Numeric PDDLGym addresses this by flattening the symbolic states and actions by instantiating all grounded fluents, functions, and actions based on the objects in the given PDDL problem. 
To motivate RL algorithms to output goal-oriented policies, we treat goal states as terminal states and define a reward function that gives a reward of one for states that achieve the goal and zero otherwise. 
Moreover, the framework supports the optional encoding of Boolean goals as binary features appended to the observation vector, allowing DRL algorithms to condition their policies on goal information when available. 

Gym environments do not support action preconditions; thus, in Numeric PDDLGym, executing an action whose preconditions are violated leaves the state unchanged. Other behaviors are possible, such as terminating the episode with a negative reward.
Also, our current conversion mechanism does not support numeric goal conditions and conditional effects.
Finally, note that while Numeric PDDLGym supports observation and action spaces of variable sizes, most DRL algorithms do not. 
Therefore, in our experiments, we limited the number of objects in the problem to a fixed size.

\begin{figure}[!t]
    \centering
    \includegraphics[width=
    0.25\textwidth]{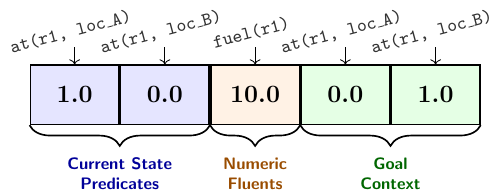}
    \caption{The Numeric PDDLGym observation encoding.}
    \label{fig:numeric-pddlgym}
\end{figure}

\noindent \textbf{Example. } Consider a simple numeric planning problem involving one robot ($r_1$) and two locations ($loc_A, loc_B$). The robot's initial state is at $loc_A$ with 10 units of fuel, and its goal is to reach $loc_B$. 
The domain defines a single lifted action, $\texttt{move}$, which moves an agent from one location to another and reduces one unit of fuel. 
Numeric PDDLGym will create two specific operators (for 
$\texttt{move}(loc_A, loc_B)$ and $\texttt{move}(loc_B, loc_A)$). Figure~\ref{fig:numeric-pddlgym} illustrates the translation from the symbolic PDDL state to the numeric observation vector.

\begin{table}[t]
\caption{Grounded statistics of the domains.}
\centering
\small
\begin{tabular}{lcc|cc} 
\hline
Domain & \multicolumn{2}{c|}{\textsc{small}} & \multicolumn{2}{c}{\textsc{large}} \\ 
       & $|obs|$ & $|action|$ & $|obs|$ & $|action|$ \\ \hline
\textbf{Counters} & 10  & 12   & 13  & 16  \\
\textbf{Sailing}  & 6  & 9   & 12   & 22  \\
\textbf{Depot}    & 62  & 256   & 80 & 450 \\
\textbf{Pogo}     & 150  & 42   & 406  & 106 \\
\hline
\end{tabular}
\label{tab:numeric-domains}
\end{table}

\begin{figure*}[t]
    \centering
    \begin{tabular}{c|c}
    \includegraphics[width=
    0.48\textwidth]{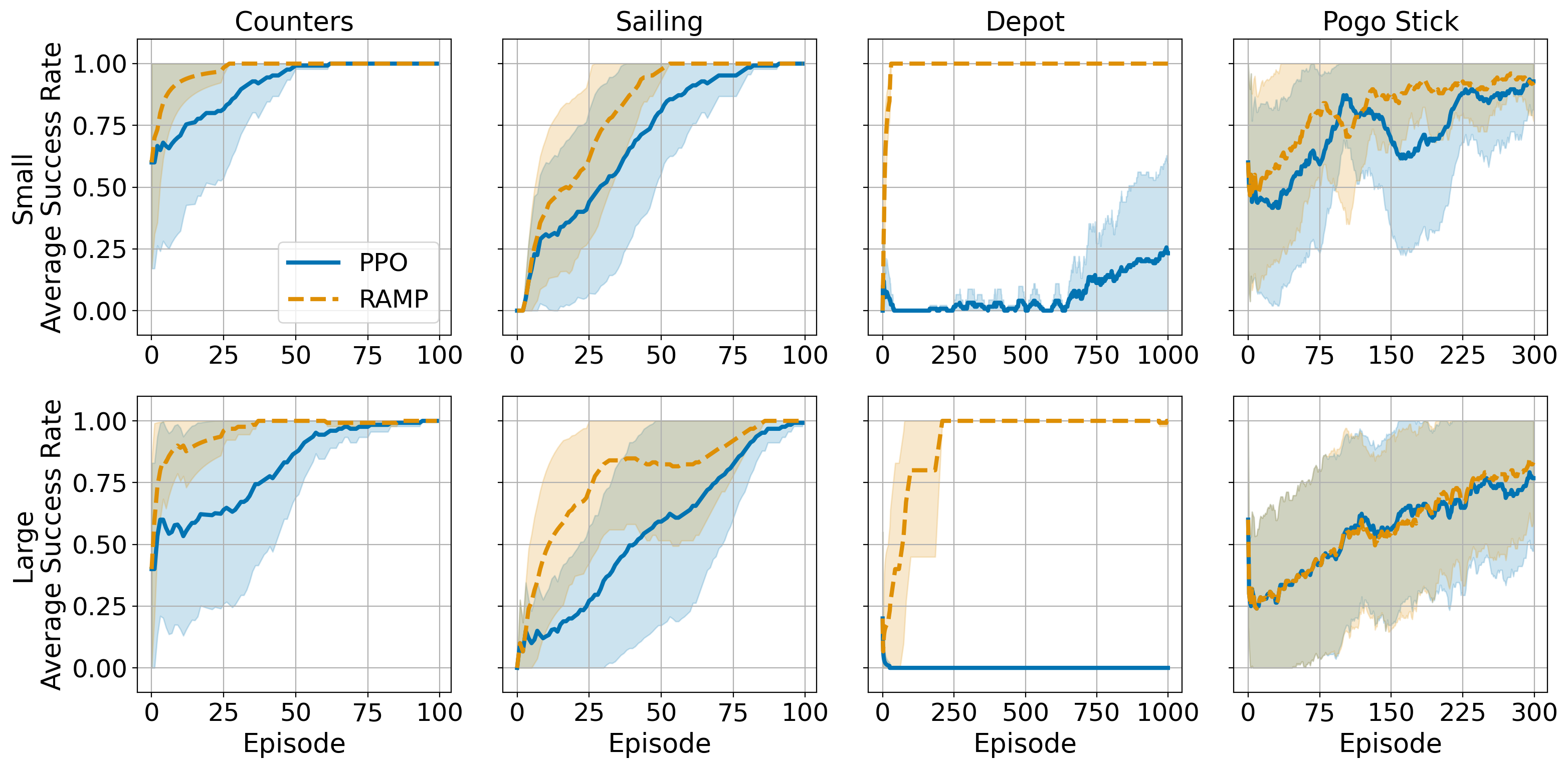}
    & 
    \includegraphics[width=
    0.48\textwidth]{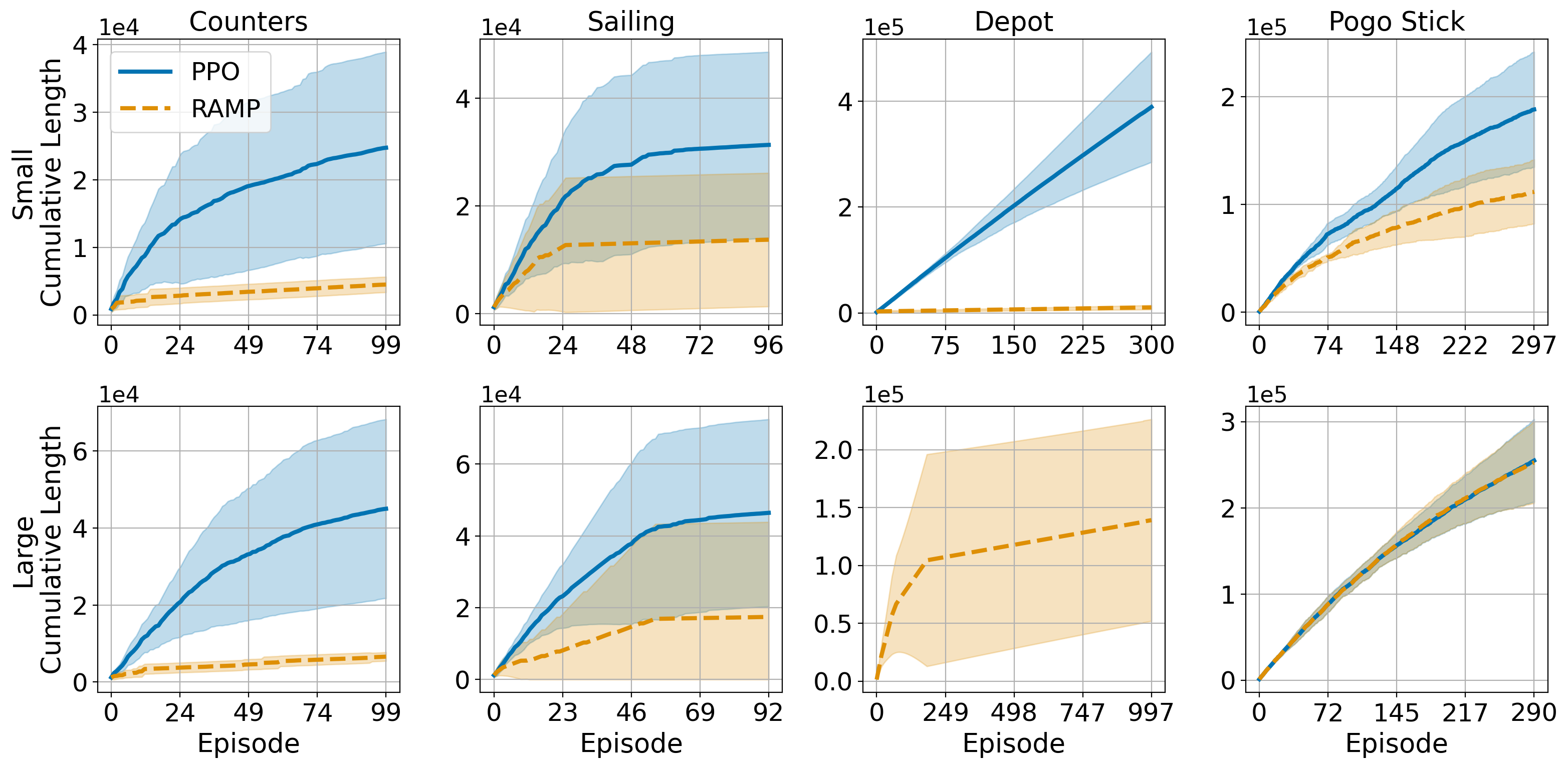}    \end{tabular}
    \caption{(Left) Rolling average success rate with 95\% confidence intervals. (Right) Cumulative solution length with 95\% confidence intervals. Top row: Small instances; bottom row: Large instances; columns: different domains.}
    \label{fig:all}
\end{figure*}

\section{Experimental Setup}

We compared RAMP against a PPO baseline, equipped with the masking technique to prevent repeating inapplicable actions.
As benchmark domains, we considered all the numeric planning domains used by \citet{mordoch2023learning}. From this set of domains, we used only domains that (1) are supported by Numeric PDDLGym (see its limitations above), (2) have linear preconditions and effects, and (3) have a problem generator. This resulted in three numeric domains:
\textsc{Counters}~\cite{scala2020subgoaling}, \textsc{Depot}~\cite{long20033rd}, and \textsc{Sailing}~\cite{scala2016heuristics}. 
To add diversity, we also experimented with \textsc{Pogo Stick}~\cite{benyamin2024crafting}, a recently introduced numeric planning domain based on Polycraft~\cite{goss2023polycraft}, a symbolic wrapper for the popular Minecraft game. 
This domain involves an agent tasked to craft a wooden pogo, which requires collecting resources of different quantities and performing several crafting recipes. See \citet{benyamin2024crafting} for more details. 
We improved their original Pogo Stick domain slightly by filtering out some inapplicable actions during problem generation to closely resemble the agent's behavior in Polycraft.

For each domain, we generated problems from two ``hardness'' levels: \emph{Small} and \emph{Large}. \emph{Large} problems have more objects, resulting in a larger set of grounded fluents, functions, and actions, which are harder to learn and to plan with. 
Table~\ref{tab:numeric-domains} reports the vector sizes of the observation and action spaces for each domain and hardness level.
We generated 50 random problem instances per domain and hardness level with a constant goal, using NSAM’s generator for IPC domains~\citep{mordoch2023learning} and the generator from~\citet{benyamin2024crafting} for Minecraft.
In each episode, a new problem instance was sampled from this set, and the evaluated algorithm was used to solve it. 
Each episode ended when either the problem had been solved (i.e., the goal had been reached) or after 1,500 environment steps. When the agent performs an illegal action (i.e., an action that does not satisfy its preconditions), it remains in place.

\noindent\textbf{Configuration and hyperparameters.} 
    For planning, we used \emph{Metric-FF} planner~\cite{hoffmann2003metric} with a 60-second time limit. 
    All experiments were repeated five times using different random seeds and conducted on a system with 16 cores of an AMD EPYC 7763 processor and 32GB of RAM.
    The \ppo hyperparameter used by the baseline \ppo and \hybrid are:
    learning rate $\alpha=10^{-3}$, 
    discount factor $\gamma = 0.999$, 
    value-function loss coefficient $0.65$, entropy coefficient $0.01$, clipping parameter $\epsilon = 0.2$, gradient clipping threshold $1.0$, and $3$ optimization epochs per update using the Adam optimizer. 
    We used a single environment runner with a batch size matched to the rollout fragment length and equal to $1500$. In each of our domains, the problems had a unique goal, and only the initial state was varied. Thus, we omitted encoding the goal in the observation space.  

\noindent\textbf{Metrics.} 
We consider two main performance metrics to assess the overall algorithm over the training set:
\emph{success rate}, computed as a moving average over the last 25 episodes; and \emph{cumulative solution length}, defined as the total number of steps performed across all successful episodes so far. 
We truncate results at the point where the success rate flattens, i.e., where further computation no longer yields meaningful improvements. 
\yarin{New:} The number of trials conducted per domain is fixed: \textsc{Counters} (100), \textsc{Sailing} (100), \textsc{Depot} (1000), and \textsc{Pogo Stick} (300).

\yarin{New:}
Moreover, we evaluate the correctness of the action model component of RAMP using the predictive power metrics proposed in \shortcite{Stern2025EvalModelLearning}, namely the precision and recall of the preconditions and Boolean effects on a test set.

\section{Results}

\miniparagraph{Solvability}
First, we consider the success rate results, depicted in Figure~\ref{fig:all}(left). 
The success rates of \hybrid and \ppo are compared across the four domains for the Small and Large hardness levels. 
As can be seen, \hybrid demonstrates superior performance in almost all cases.
In Counters and Sailing, \hybrid reaches near-perfect solvability much faster than \ppo.
In Depot, \hybrid maintains a clear advantage, solving problems even in the hard instances, where PPO could not solve any of the problems.
In these problems, the planning component in \hybrid is able to find a solution with the learned domain model in over 90\% of cases, leading to higher overall solvability rates. 
The results for the Pogo Stick domain are less conclusive: in the small hardness level \hybrid still outperforms \ppo, but for the large hardness level, there is no statistically significant difference. 
We conjecture that this is because the planner is limited to 60 seconds, and thus it may fail to find a solution even when one exists. Indeed, the success rate drops to around 60\% for small Pogo Stick instances and approaches zero for large instances, which leads to \hybrid and \ppo achieving comparable success rates.

\miniparagraph{Plan Quality}
Next, consider the \emph{cumulative solution length} results in 
Figure~\ref{fig:all}(right). Results are in log scale, where lower values indicate better performance (and shorter plans).
For the Large instances in Depot, \hybrid was the only agent to solve any instance; therefore, only its results are reported.
\hybrid consistently finds significantly shorter plans than \ppo across both difficulty levels, except in the pogo task, where plan lengths are similar due to the planner frequently failing on large instances, limiting RAMP’s ability to optimize the plan.
This highlights the benefit of the planner in guiding the RL agent toward more efficient solutions, rather than just finding \emph{any} solution.
Even when the planner does not always succeed in finding a plan, its guidance improves the quality and efficiency of the solutions that are found.
\yarin{New part:}
Although there is a theoretical risk that the RL agent may inherit sub-optimality from an incomplete or suboptimal planner, we did not observe this in practice.

\miniparagraph{Action Model Quality}
To evaluate the action model part of RAMP, we computed the average precision and recall of the learned preconditions and effects against the ground-truth domain models. We ran Metric-FF and 200 random walks on 50 newly generated instances to estimate these metrics. For action effects, RAMP achieves perfect performance, with precision and recall of $1.0$ across all evaluated domains. For preconditions, the safety guarantees of NSAM ensure a precision of $1.0$.

Recall varies based on the exploration data gathered (Table~\ref{tab:combined_results} train columns). 
Counters achieves near-perfect recall, while Depot remains significantly lower, and Sailing and Pogo Stick fall in between. 
Compared to the standard offline NSAM trained on an expert dataset of 80 trajectories, the offline approach achieves perfect recall, whereas our method does not.
This is because our algorithm prioritizes task solvability over exhaustive recovery of the full action model. 
For example, in the Depot (small) domain, only 4.2 trajectories on average were sufficient to learn a model capable of solving subsequent problems, indicating that perfect recall is not essential for effective planning.

\begin{table}[t]
\caption{Performance of NSAM in RAMP across domains and sizes. 
Precondition recall is measured on a held-out test set, while solved and timeout statistics are collected during online training.}
\centering
\begin{tabular}{|l|l|c|cc|}
\hline
 &  & \multicolumn{1}{c|}{\textbf{Test}} & \multicolumn{2}{c|}{\textbf{Train}} \\
\textbf{Domain} & \textbf{Size} & \textbf{Recall} & \textbf{Solved} & \textbf{Timeout} \\
\hline
counters    & small & 0.960 & 97.4  & 1.2 \\
counters    & large & 0.990 & 96.6  & 2.0 \\
\hline
sailing     & small & 0.696 & 87.4  & 0.8 \\
sailing     & large & 0.849 & 85.2  & 7.0 \\
\hline
depot       & small & 0.128 & 995.8 & 2.6 \\
depot       & large & 0.217 & 931.8 & 60.4 \\
\hline
pogo\_stick & small & 0.677 & 169.8 & 109.8 \\
pogo\_stick & large & 0.729 & 0.4   & 281.8 \\
\hline
\end{tabular}
\label{tab:combined_results}
\end{table}

\miniparagraph{Plan Efficiency}
The DRL policy actively leverages the planner whenever a valid plan exists; across our training instances (as seen in Table~\ref{tab:combined_results} test column), RAMP successfully utilizes the planner's plan in over 85\% of cases for Counters and Sailing, and in over 93\% of the cases for Depot. The planner is only ignored when it fails to find a solution (e.g., timing out on Large Pogo Stick instances), at which point PPO defaults to independent exploration. This ensures that the agent consistently benefits from the high-quality, efficient plans produced by the symbolic model when available, without inheriting the planner's incompleteness.

\section{Conclusions and Future Work}
In this work, we introduced \hybrid, a hybrid strategy for online numeric planning that integrates RL, numeric action model learning (\nsam), and planning. 
By simultaneously learning a safe domain model and an RL policy, \hybrid creates a positive feedback loop: the model generates plans to guide the agent, while the agent's exploration refines the model.
Experiments show \hybrid significantly outperforms PPO in both solvability and solution length on 3 IPC domains and a recently introduced Minecraft domain. 

\yarin{added:}
Moreover, we introduced Numeric PDDLGym, a framework that converts PDDL2.1 domains into Gym environments with fixed-size observation and action spaces. By grounding the symbolic representations and converting them into fixed-length vectors, it enables the direct application of standard RL and DRL methods to numeric planning domains.

Future work will focus on relaxing the assumption of noise-free observability by incorporating probabilistic state representations and noise-robust action model learning, enabling deployment in realistic, partially observable environments.



\yarin{new:}
\begin{acks}

Yarin Benyamin gratefully acknowledges support from the STEM Scholarship for Outstanding Doctoral Students at the Kreitman School of Advanced Graduate Studies.
This research was funded by ISF grants No. 1238/23 to Roni Stern. Additional support was provided by Israel’s Ministry of Innovation, Science and Technology (MOST) under Grant No. 1001706842, in collaboration with the Israel National Road Safety Authority and Netivei Israel, awarded to Shahaf Shperberg.

\end{acks}



\bibliographystyle{ACM-Reference-Format} 
\bibliography{sample}

@article{aineto2019learning,
  title={Learning action models with minimal observability},
  author={Aineto, Diego and Celorrio, Sergio Jim{\'e}nez and Onaindia, Eva},
  journal={Artificial Intelligence},
  volume={275},
  pages={104--137},
  year={2019},
  publisher={Elsevier}
}

@inproceedings{stern2017efficient,
  title={Efficient, Safe, and Probably Approximately Complete Learning of Action Models},
  author={Roni Stern and Brendan Juba},
booktitle={the International Joint Conference on Artificial Intelligence (IJCAI)},
 pages = {4405--4411},
  year={2017}
}

@inproceedings{juba2021safe,
  author    = {Brendan Juba and
               Hai S. Le and
               Roni Stern},
  title     = {Safe Learning of Lifted Action Models},
  booktitle = {International Conference on Principles of
               Knowledge Representation and Reasoning ({KR})},
  pages     = {379--389},
  year      = {2021}
}

@article{goss2023polycraft,
  title={Polycraft World AI Lab (PAL): An Extensible Platform for Evaluating Artificial Intelligence Agents},
  author={Goss, Stephen A and Steininger, Robert J and Narayanan, Dhruv and Oliven{\c{c}}a, Daniel V and Sun, Yutong and Qiu, Peng and Amato, Jim and Voit, Eberhard O and Voit, Walter E and Kildebeck, Eric J},
  journal={arXiv preprint arXiv:2301.11891},
  year={2023}
}

@article{schulman2017proximal,
  title={Proximal policy optimization algorithms},
  author={Schulman, John and Wolski, Filip and Dhariwal, Prafulla and Radford, Alec and Klimov, Oleg},
  journal={arXiv preprint arXiv:1707.06347},
  year={2017}
}

@article{fox2003pddl2,
  title={PDDL2. 1: An extension to PDDL for expressing temporal planning domains},
  author={Fox, Maria and Long, Derek},
  journal={Journal of artificial intelligence research},
  volume={20},
  pages={61--124},
  year={2003}
}

@inproceedings{mordoch2023learning,
  author       = {Argaman Mordoch and
                  Brendan Juba and
                  Roni Stern},
  title        = {Learning Safe Numeric Action Models},
  booktitle    = {{AAAI}},
  pages        = {12079--12086},
  publisher    = {{AAAI} Press},
  year         = {2023}
}

@book{sutton2018reinforcement,
  title={Reinforcement learning: An introduction},
  author={Sutton, Richard S and Barto, Andrew G},
  year={2018},
  publisher={MIT press}
}

@phdthesis{watkins1989learning,
author={Watkins, Christopher John Cornish Hellaby},
title={Learning from delayed rewards},
  publisher={King's College, Cambridge United Kingdom},
school={Oxford: King's College},
  year={1989},
}

@article{watkins1992q,
  title={Q-learning},
  author={Watkins, Christopher JCH and Dayan, Peter},
  journal={Machine learning},
  volume={8},
  pages={279--292},
  year={1992},
  publisher={Springer}
}

@book{ghallab2004automated,
  title={Automated Planning: theory and practice},
  author={Ghallab, Malik and Nau, Dana and Traverso, Paolo},
  year={2004}
}

@article{hoffmann2003metric,
  title={The Metric-FF Planning System: Translating ``Ignoring Delete Lists'' to Numeric State Variables},
  author={Hoffmann, J{\"o}rg},
  journal={Journal of Artificial Intelligence Research},
  volume={20},
  pages={291--341},
  year={2003}
}

@article{cresswell2013acquiring,
  title={Acquiring planning domain models using LOCM},
  author={Cresswell, Stephen and McCluskey, Thomas and West, Margaret},
  journal={The Knowledge Engineering Review},
  volume={28},
  number={2},
  pages={195--213},
  year={2013}
}

@article{segura2021discovering,
  title={Discovering relational and numerical expressions from plan traces for learning action models},
  author={Segura-Muros, Jos{\'e} {\'A} and P{\'e}rez, Ra{\'u}l and Fern{\'a}ndez-Olivares, Juan},
  journal={Applied Intelligence},
  pages={1--17},
  year={2021},
  publisher={Springer}
}

@inproceedings{mordoch2024-nsam-star,
  author       = {Argaman Mordoch and
                  Shahaf S. Shperberg and
                  Roni Stern and
                  Brendan Juba},
  title        = {Enhancing Numeric-SAM for Learning with Few Observations},
  booktitle      = {ICAPS Workshop on Knowledge Engineering for Planning and Scheduling (KEPS)},
  year         = {2024}
}

@article{sreedharan2023optimistic,
  title={Optimistic exploration in reinforcement learning using symbolic model estimates},
  author={Sreedharan, Sarath and Katz, Michael},
  journal={Advances in Neural Information Processing Systems},
  volume={36},
  pages={34519--34535},
  year={2023}
}

@inproceedings{Lamanna24,
  author       = {Leonardo Lamanna and
                  Luciano Serafini},
  title        = {Action Model Learning from Noisy Traces: a Probabilistic Approach},
  booktitle    = {{ICAPS}},
  pages        = {342--350},
  publisher    = {{AAAI} Press},
  year         = {2024}
}

@inproceedings{MordochSSJ24,
  author       = {Argaman Mordoch and
                  Enrico Scala and
                  Roni Stern and
                  Brendan Juba},
  title        = {Safe Learning of {PDDL} Domains with Conditional Effects},
  booktitle    = {{ICAPS}},
  pages        = {387--395},
  publisher    = {{AAAI} Press},
  year         = {2024}
}

@inproceedings{jin2022creativity,
  title={Creativity of {AI}: Automatic symbolic option discovery for facilitating deep reinforcement learning},
  author={Jin, Mu and Ma, Zhihao and Jin, Kebing and Zhuo, Hankz Hankui and Chen, Chen and Yu, Chao},
  booktitle={Proceedings of the AAAI Conference on Artificial Intelligence},
  volume={36},
  number={6},
  pages={7042--7050},
  year={2022}
}

@inproceedings{chitnis2021glib,
  title={Glib: Efficient exploration for relational model-based reinforcement learning via goal-literal babbling},
  author={Chitnis, Rohan and Silver, Tom and Tenenbaum, Joshua B and Kaelbling, Leslie Pack and Lozano-P{\'e}rez, Tom{\'a}s},
  booktitle={Proceedings of the AAAI Conference on Artificial Intelligence},
  volume={35},
  number={13},
  pages={11782--11791},
  year={2021}
}

@inproceedings{ng2019incremental,
  title={Incremental Learning of Planning Actions in Model-Based Reinforcement Learning},
  author={Ng, Jun Hao Alvin and Petrick, Ronald PA},
  booktitle={IJCAI},
  pages={3195--3201},
  year={2019}
}

@inproceedings{karia2023epistemic,
  title={Epistemic Exploration for Generalizable Planning and Learning in Non-Stationary Stochastic Settings},
  author={Karia, Rushang and Verma, Pulkit and Vipat, Gaurav and Srivastava, Siddharth},
  booktitle={NeurIPS 2023 Workshop on Generalization in Planning},
  year={2023}
}

@article{verma2023autonomous,
  title={Autonomous capability assessment of sequential decision-making systems in stochastic settings},
  author={Verma, Pulkit and Karia, Rushang and Srivastava, Siddharth},
  journal={Advances in Neural Information Processing Systems},
  volume={36},
  pages={54727--54739},
  year={2023}
}

@inproceedings{le2024learning,
  title={Learning Safe Action Models with Partial Observability},
  author={Le, Hai S and Juba, Brendan and Stern, Roni},
  booktitle={Proceedings of the AAAI Conference on Artificial Intelligence},
  volume={38},
  number={18},
  pages={20159--20167},
  year={2024}
}

@inproceedings{juba2022learning,
  title={Learning probably approximately complete and safe action models for stochastic worlds},
  author={Juba, Brendan and Stern, Roni},
  booktitle={Proceedings of the AAAI Conference on Artificial Intelligence},
  volume={36},
  number={9},
  pages={9795--9804},
  year={2022}
}

@inproceedings{lamanna2021online,
  title={Online Learning of Action Models for PDDL Planning.},
  author={Lamanna, Leonardo and Saetti, Alessandro and Serafini, Luciano and Gerevini, Alfonso and Traverso, Paolo and others},
  booktitle={IJCAI},
  pages={4112--4118},
  year={2021}
}

@inproceedings{asai2018classical,
  title={Classical planning in deep latent space: Bridging the subsymbolic-symbolic boundary},
  author={Asai, Masataro and Fukunaga, Alex},
  booktitle={Proceedings of the aaai conference on artificial intelligence},
  volume={32},
  number={1},
  year={2018}
}

@inproceedings{xi2024neuro,
  title={Neuro-Symbolic Learning of Lifted Action Models from Visual Traces},
  author={Xi, Kai and Gould, Stephen and Thi{\'e}baux, Sylvie},
  booktitle={Proceedings of the International Conference on Automated Planning and Scheduling},
  volume={34},
  pages={653--662},
  year={2024}
}

@inproceedings{jang2017categorical,
  title={Categorical Reparametrization with Gumble-Softmax},
  author={Jang, Eric and Gu, Shixiang and Poole, Ben},
  booktitle={International Conference on Learning Representations (ICLR)},
  year={2017}
}

@inproceedings{james2022autonomous,
  title={Autonomous learning of object-centric abstractions for high-level planning},
  author={James, Steven and Rosman, Benjamin and Konidaris, GD},
  booktitle={International Conference on Learning Representations (ICLR)},
  year={2022}
}

@inproceedings{benyamin2024crafting,
  title={Crafting a Pogo Stick in Minecraft with Heuristic Search},
  author={Benyamin, Yarin and Mordoch, Argaman and Shperberg, Shahaf and Piotrowski, Wiktor and Stern, Roni},
  booktitle={International Symposium on Combinatorial Search},
  pages={261--262},
  year={2024}
}

@inproceedings{HuangO22,
  author       = {Shengyi Huang and
                  Santiago Onta{\~{n}}{\'{o}}n},
  title        = {A Closer Look at Invalid Action Masking in Policy Gradient Algorithms},
  booktitle    = {{FLAIRS}},
  year         = {2022}
}

@article{long20033rd,
  title={The 3rd international planning competition: Results and analysis},
  author={Long, Derek and Fox, Maria},
  journal={Journal of Artificial Intelligence Research},
  volume={20},
  pages={1--59},
  year={2003}
}

@inproceedings{scala2016heuristics,
  title={Heuristics for Numeric Planning via Subgoaling},
  author={Scala, Enrico and Haslum, Patrik and Thi{\'e}baux, Sylvie},
  booktitle={IJCAI},
  pages={3228--3234},
  year={2016}
}

@article{scala2020subgoaling,
  title={Subgoaling techniques for satisficing and optimal numeric planning},
  author={Scala, Enrico and Haslum, Patrik and Thi{\'e}baux, Sylvie and Ramirez, Miquel},
  journal={Journal of Artificial Intelligence Research},
  volume={68},
  pages={691--752},
  year={2020}
}

@article{towers2024gymnasium,
  title={Gymnasium: A Standard Interface for Reinforcement Learning Environments},
  author={Towers, Mark and Kwiatkowski, Ariel and Terry, Jordan and Balis, John U and De Cola, Gianluca and Deleu, Tristan and Goul{\~a}o, Manuel and Kallinteris, Andreas and Krimmel, Markus and KG, Arjun and others},
  journal={arXiv preprint arXiv:2407.17032},
  year={2024}
}

@article{silver2020pddlgym,
  title={PDDLGym: Gym environments from PDDL problems},
  author={Silver, Tom and Chitnis, Rohan},
  journal={arXiv preprint arXiv:2002.06432},
  year={2020}
}

@article{vinyals2017starcraft,
  title={StarCraft II: A new challenge for reinforcement learning},
  author={Vinyals, Oriol and Ewalds, Timo and Bartunov, Sergey and Georgiev, P and Vezhnevets, Alexander S and Yeo, Michelle and Makhzani, Alireza and K{\"u}ttler, Heinrich and Agapiou, John and Schrittwieser, Julian and others},
  journal={arXiv preprint arXiv:1708.04782},
  year={2017}
}

@article{berner2019dota,
  title={Dota 2 with large scale deep reinforcement learning},
  author={Berner, Christopher and Brockman, Greg and Chan, Brooke and Cheung, Vicki and Dkebiak, Przemys{\l}aw and Dennison, Christy and Farhi, David and Fischer, Quirin and Hashme, Shariq and Hesse, Chris and J{\'o}zefowicz, Rafa{\l} and Gray, Scott and Olsson, Catherine and Pachocki, Jakub and Petrov, Michael and de Oliveira Pinto, Henrique Ponda and Raiman, Jonathan and Salimans, Tim and Schlatter, Jeremy and Schneider, Jonas and Sidor, Szymon and Sutskever, Ilya and Tang, Jie and Wolski, Filip and Zhang, Susan},
  journal={arXiv preprint arXiv:1912.06680},
  year={2019}
}

@inproceedings{ye2020mastering,
  title={Mastering complex control in {MOBA} games with deep reinforcement learning},
  author={Ye, Deheng and Liu, Zhao and Sun, Mingfei and Shi, Bei and Zhao, Peilin and Wu, Hao and Yu, Hongsheng and Yang, Shaojie and Wu, Xipeng and Guo, Qingwei and Chen, Qiaobo and Yin, Yinyuting and Zhang, Hao and Shi, Tengfei and Wang, Liang and Fu, Qiang and Yang, Wei and Huang, Lanxiao},
  booktitle={AAAI},
  volume={34},
  number={04},
  pages={6672--6679},
  year={2020},
  doi={10.1609/aaai.v34i04.6144}
}

@article{stable-baselines3,
  author  = {Antonin Raffin and Ashley Hill and Adam Gleave and Anssi Kanervisto and Maximilian Ernestus and Noah Dormann},
  title   = {Stable-Baselines3: Reliable Reinforcement Learning Implementations},
  journal = {Journal of Machine Learning Research},
  year    = {2021},
  volume  = {22},
  number  = {268},
  pages   = {1-8},
  url     = {http://jmlr.org/papers/v22/20-1364.html}
}

@inproceedings{liang2018rllib,
    title={{RLlib}: Abstractions for Distributed Reinforcement Learning},
    author={
        Eric Liang and
        Richard Liaw and
        Robert Nishihara and
        Philipp Moritz and
        Roy Fox and
        Ken Goldberg and
        Joseph E. Gonzalez and
        Michael I. Jordan and
        Ion Stoica
    },
    booktitle = {International Conference on Machine Learning ({ICML})},
    year={2018},
    url={https://arxiv.org/pdf/1712.09381}
}

@InProceedings{Stern2025EvalModelLearning,
  author    = {Roni Stern and Leonardo Lamanna and Argaman Mordoch and Yarin Benyamin and Pascal Lauer and Brendan Juba and Gregor Behnke and Christian Muise and Pascal Bercher and Mauro Vallati and Kai Xi and Omar Wattad and Omer Eliyahu},
  title     = {Evaluating Planning Model Learning Algorithms},
  booktitle = {Workshop on Knowledge Engineering for Planning and Scheduling (KEPS) at ICAPS},
  year ={2025},
}

@inproceedings{hessel2018rainbow,
  title={Rainbow: Combining improvements in deep reinforcement learning},
  author={Hessel, Matteo and Modayil, Joseph and Van Hasselt, Hado and Schaul, Tom and Ostrovski, Georg and Dabney, Will and Horgan, Dan and Piot, Bilal and Azar, Mohammad and Silver, David},
  booktitle={Proceedings of the AAAI conference on artificial intelligence},
  volume={32},
  number={1},
  year={2018}
}

@inproceedings{haarnoja2018soft,
  title={Soft actor-critic: Off-policy maximum entropy deep reinforcement learning with a stochastic actor},
  author={Haarnoja, Tuomas and Zhou, Aurick and Abbeel, Pieter and Levine, Sergey},
  booktitle={International conference on machine learning},
  pages={1861--1870},
  year={2018},
  organization={Pmlr}
}

@article{younes2004ppddl1,
  title={PPDDL1. 0: An extension to PDDL for expressing planning domains with probabilistic effects},
  author={Younes, H{\aa}kan LS and Littman, Michael L},
  journal={Techn. Rep. CMU-CS-04-162},
  volume={2},
  number={99},
  pages={12},
  year={2004}
}

@inproceedings{tantakoun2025llms,
  title={LLMs as planning formalizers: A survey for leveraging large language models to construct automated planning models},
  author={Tantakoun, Marcus and Muise, Christian and Zhu, Xiaodan},
  booktitle={Findings of the Association for Computational Linguistics: ACL 2025},
  pages={25167--25188},
  year={2025}
}

@article{benyamin2025toward,
  title={Toward PDDL Planning Copilot},
  author={Benyamin, Yarin and Mordoch, Argaman and Shperberg, Shahaf S and Stern, Roni},
  journal={arXiv preprint arXiv:2509.12987},
  year={2025}
}

@inproceedings{lindsay2017framer,
  title={Framer: Planning Models from Natural Language Action Descriptions},
  author={Lindsay, Alan and Read, Jonathon and Ferreira, Joao F and Hayton, Thomas and Porteous, Julie and Gregory, PJ},
  booktitle={International Conference on Automated Planning and Scheduling (ICAPS)},
  year={2017}
}


\end{document}